\documentclass[10pt,twocolumn,letterpaper]{article}

 \usepackage{iccv}              

\definecolor{iccvblue}{rgb}{0.21,0.49,0.74}
\usepackage[pagebackref,breaklinks,colorlinks,allcolors=iccvblue]{hyperref}
\graphicspath{{./images/}}

\def\paperID{*****} 
\def\confName{ICCV}
\def\confYear{2025}

\renewcommand{\v}[1]{\boldsymbol{#1}}
\newcommand{\vH}[1]{\textbf{#1}}
\newcommand{\mat}[1]{\v{#1}}
\newcommand{\matH}[1]{\vH{#1}}
\newcommand{\vN}[1]{\bar{\v{#1}}}

\title{A Calibration Tool for Refractive Underwater Vision}

\author{Felix Seegräber$^{\text{1}}$ \hspace{0.7em} \and Mengkun She$^{\text{1}}$ \hspace{0.7em} \and Felix Woelk$^{\text{2}}$ \hspace{0.7em} \and Kevin Köser$^{\text{1}}$
\and $^{\text{1}}$Kiel University \and $^{\text{2}}$University of Applied Sciences Kiel
}

\begin{document}

\maketitle
\thispagestyle{empty}
\pagestyle{empty}

\begin{abstract}
Many underwater applications rely on vision sensors and require proper camera calibration, i.e. knowing the incoming light ray for each pixel in the image. While for the ideal pinhole camera model all viewing rays intersect in a single 3D point, underwater cameras suffer from - possibly multiple - refractions of light rays at the interfaces of water, glass and air. These changes of direction depend on the position and orientation of the camera inside the water-proof housing, as well as on the shape and properties of the optical window, the port, itself.
In recent years explicit models for underwater vision behind common ports such as flat or dome port have been proposed, but the underwater community is still lacking a calibration tool which can determine port parameters through refractive calibration. With this work we provide the first open source implementation of an underwater refractive camera calibration toolbox.
It allows end-to-end calibration of underwater vision systems, including camera, stereo and housing calibration for systems with dome or flat ports. The implementation is verified using rendered datasets and real-world experiments.
\end{abstract}

\section{Introduction}

Camera calibration is a prerequisite for many computer vision applications. Relating image content to real world objects and measurements requires knowledge about the image formation characteristics of each utilized camera. The parameters defining this image formation, typically focal length, principal point and lens distortion parameters, are estimated during calibration. When acquiring images underwater, the light must travel between water, glass and air, media with different refractive indices, causing the light to be refracted at the interfaces. For certain use cases, e.g. keeping the camera at fixed working distances from objects, the refractive distortion can be absorbed by the lens distortion parameters and focal length of conventional camera models if the whole setup is calibrated underwater \cite{Lavest2000, Shortis2015}. In the general case, and in particular for settings where objects are observed from varying viewing distances, the pinhole approximation can introduce significant systematic errors, since the amount of geometric distortion depends on scene distance \cite{Sedlazeck2012, Treibitz2012, kahmen2020impact} and on the interface to projection center distance for flat ports \cite{Luczynski2017, nocerino2021bundle}. Additionally refraction can cause other image degradation effects such as chromatic aberration \cite{menna2017optical}.

\begin{figure}[t!]
	\centering
	\includegraphics[width=0.9\columnwidth,trim={0 5cm 17.5cm 0},clip]{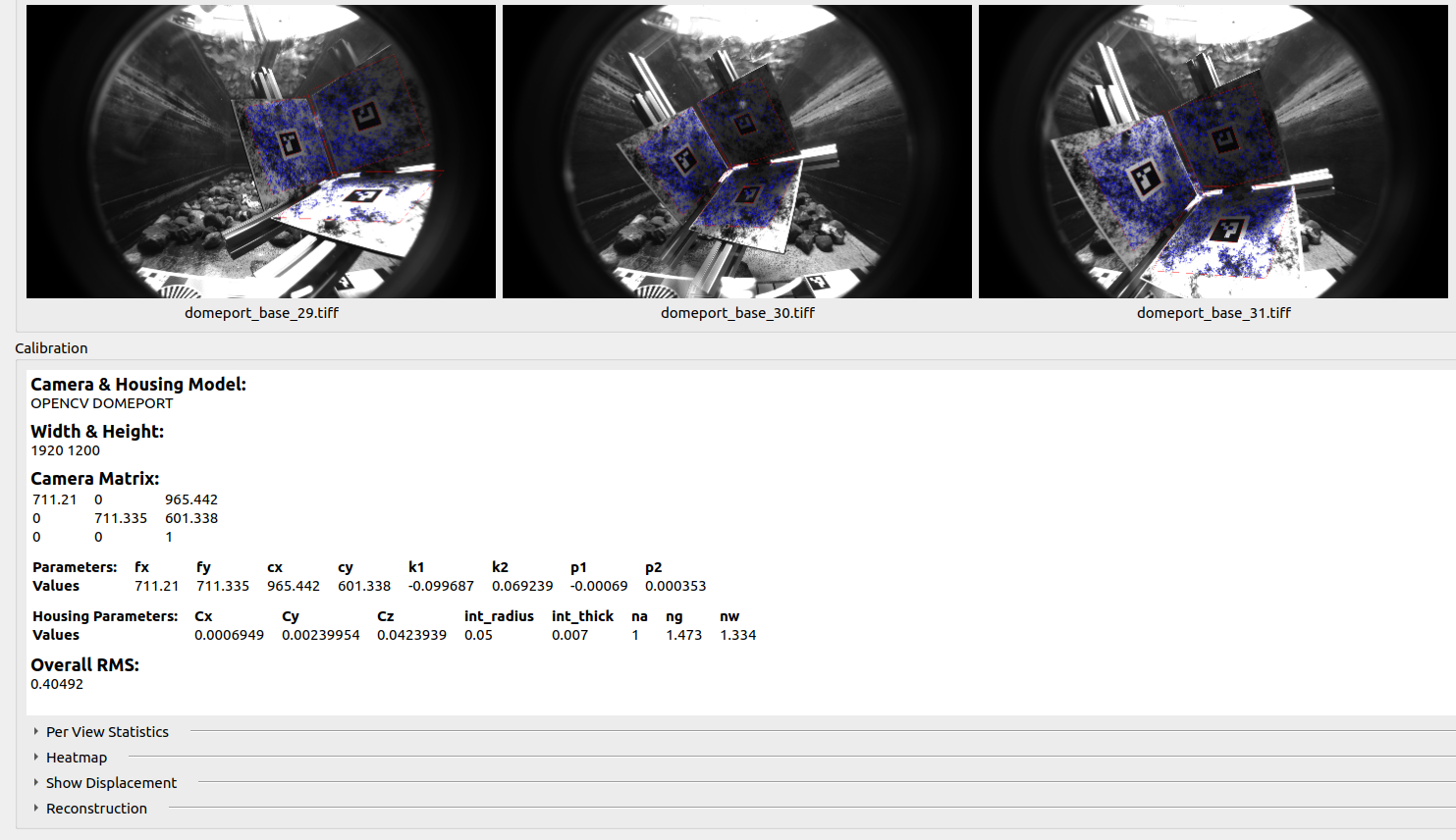}
	\includegraphics[width=0.9\columnwidth]{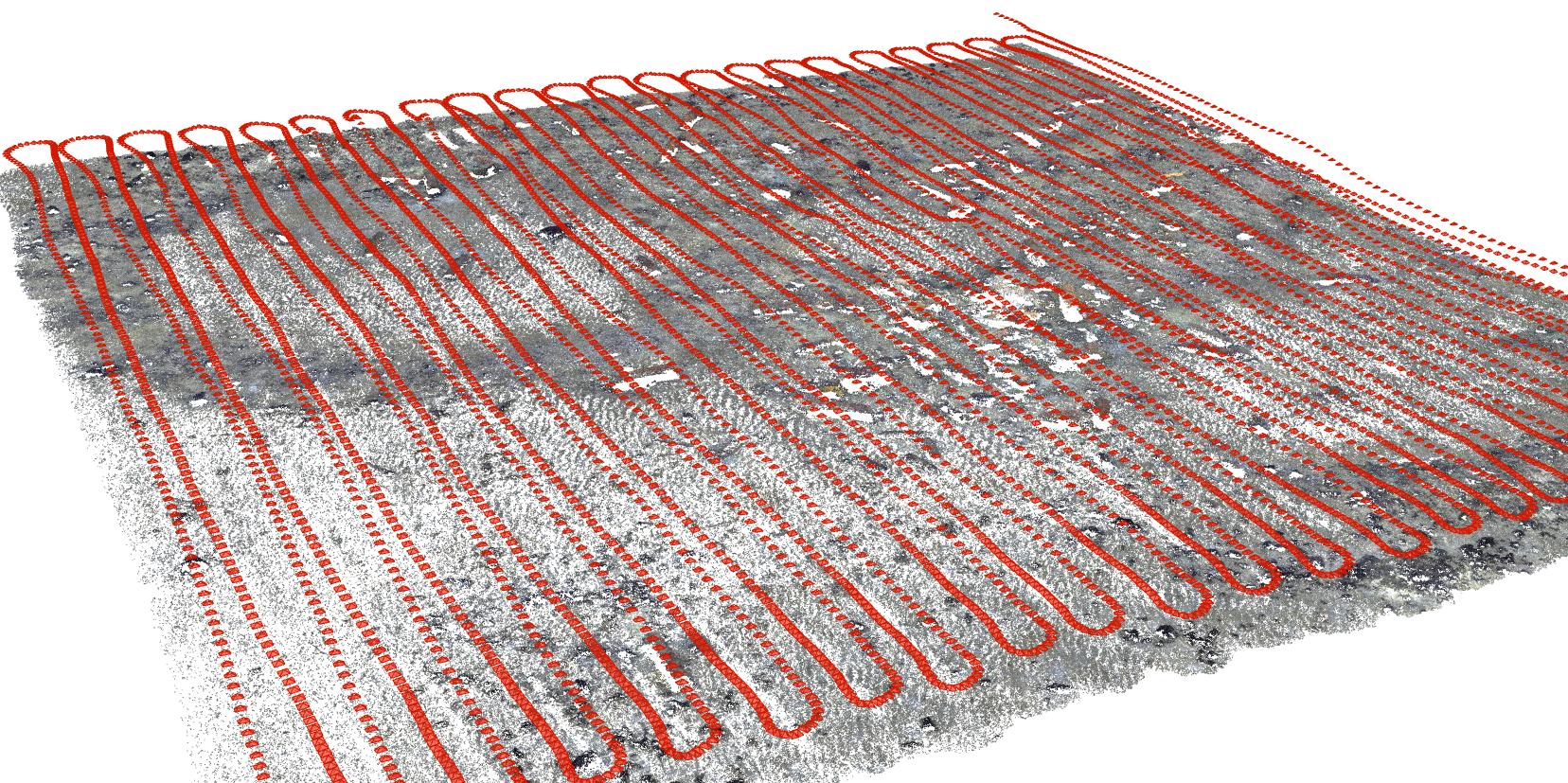}
	\caption{Crop of the graphical user interface (\textit{top}) and an example of an application explicitly modeling underwater housings for refractive Structure-From-Motion (\textit{bottom}) from \cite{She2022}.} \label{fig:intro}
\end{figure}

An alternative approach for underwater cameras is to therefore not only model the camera, but also the housing protecting it. Theory for ray-tracing through refractive interfaces has been described \cite{kotowski1988phototriangulation, Glassner1989, maas1992digitale} and solutions for calibrating the two main setups, housings with a flat glass pane '\textit{Flat Port}' and a dome shaped one '\textit{Dome Port}' have been developed in e.g. \cite{JordtSedlazeck2012, Agrawal2012,elnashef2023theory} and \cite{she2019adjustment,She2022, Kunz2008}. Several underwater applications use these explicit housing models i.e. in Structure-From-Motion \cite{JordtSedlazeck2013,jordt2016refractive,Chadebecq2019,Kang2017,Kang2012}, Multi-View-Stereo 3D reconstruction \cite{Seegraeber2023} or visual odometry \cite{Elnashef2023}. 
These approaches require knowledge about the housing parameters, which are hard to accurately measure, but have great influence on the image formation. 
Even methods aiming to compensate refraction such as \cite{Luczynski2017} or those that use a simplified model \cite{Kang2017} either need to know the flat port normal or assume perfect alignment. 
Similarly, dome port housings avoid refraction when the entrance pupil of the lens is perfectly centered inside the dome. 
In both cases it is useful to verify the alignment of housing and camera, either to model it explicitly or to assure that the assembly is correct. Housing calibration allows to precisely estimate the relevant properties. 
Despite these advantages, and despite the fact that several approaches have been proposed which explicitly model the housing, there are no available implementations or toolboxes for underwater camera calibration to the best of our knowledge. 
Our proposed application is meant as a service to the community and aims to fill this gap. We provide a cross platform, stand-alone tool which allows straight forward camera, stereo and housing calibration for users with varying technical background. The application is provided as an open-source implementation\footnote{\url{https://cau-git.rz.uni-kiel.de/inf-ag-koeser/calibmar}}. \smallskip
\\
\textbf{Contributions} We present a calibration tool for underwater housing calibration with the main contribution being the release of the application and source code for the scientific community. The main capabilities and features are listed below:
\begin{itemize}
	\item \textbf{Camera calibration} with varying degrees of distortion model complexity, abstracted as predefined camera models.
	\item \textbf{Housing calibration} for dome and flat ports given the intrinsics of one of the available camera models.
	\item \textbf{Stereo calibration}, with or without optimizing camera intrinsics.
	\item \textbf{Auto-calibration} for camera or housing from SIFT features, optionally using a novel masking target to provide more control over the calibration.
	\item \textbf{Live calibration} for connected USB cameras including an implementation of \textit{Calibration Wizard} \cite{Peng2018} and several other convenience and quality assessment features.
\end{itemize}

\section{Previous Work}

For regular in-air calibration several tools exist. The MATLAB toolbox \cite{Bouguet2022} and OpenCV library \cite{OpenCV2024} are commonly known examples. Next to a few proprietary, commercial offerings some open-source specialized solutions are available as well. There are tools focusing on multi-camera systems \cite{Furgale2013, Li2013}, combining camera and IMU calibration \cite{Furgale2013} for robotics applications or calibrating omnidirectional \cite{Scaramuzza2006} and multi-spectral cameras \cite{Schramm2021}. 
The prevalence of these tools shows the importance of camera calibration for many fields. While, as mentioned in the previous section, various underwater vision approaches explicitly require the camera housings and theory for their calibration exists, no tools for this use case have been published yet, which is the motivation for our work that is outlined in the next sections.

\begin{figure}[h]
	\centering
	\includegraphics[width=0.4\columnwidth]{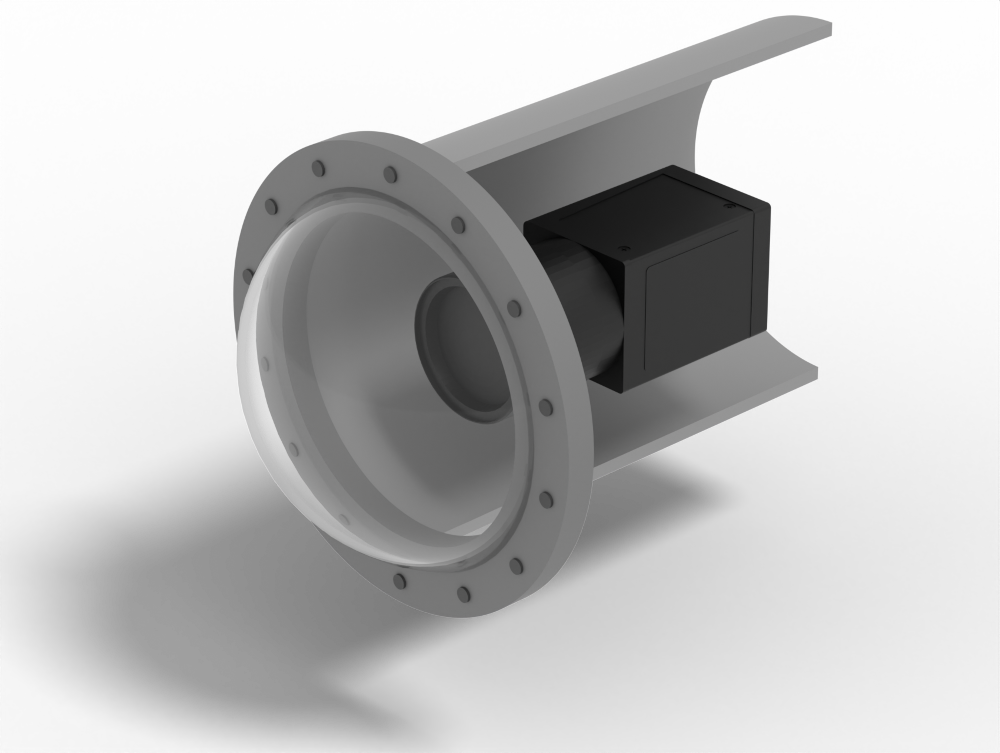}
	\includegraphics[width=0.4\columnwidth]{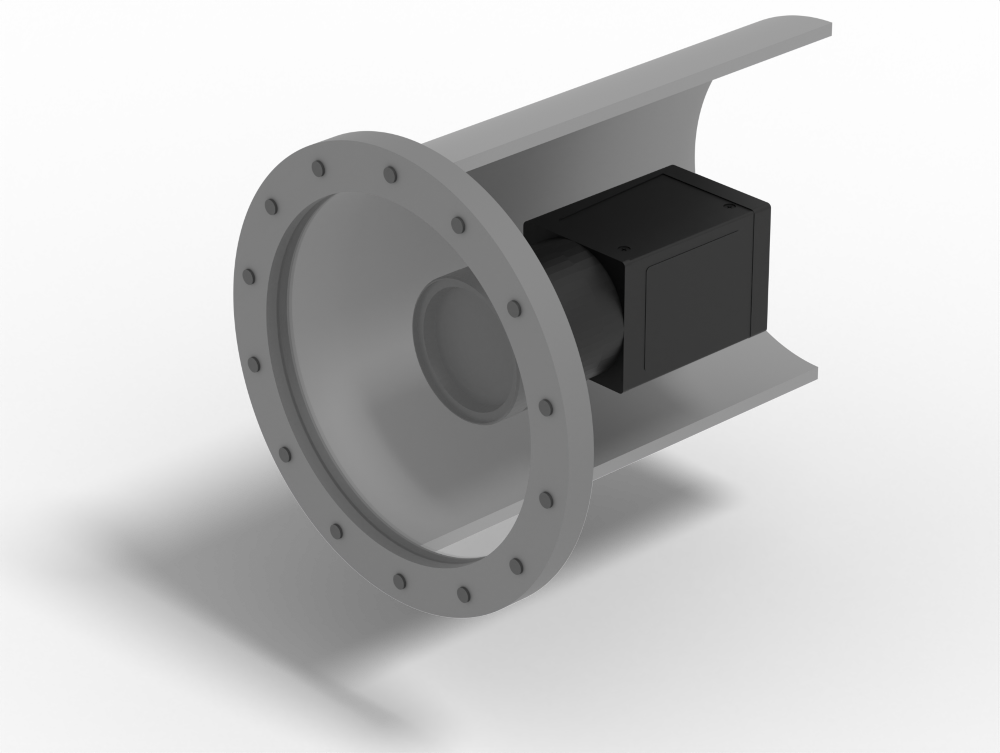}
	\caption{Schematic dome and flat port housings} \label{fig:housings}
\end{figure}

\section{Application Structure and Features}

This section introduces the main features and gives a short introduction into the underlying theory and algorithms used. The three main use cases are camera, housing and stereo calibration. All three features are frequently needed in underwater computer vision applications. The tool is implemented as a stand-alone, cross-platform C++ application with a graphical user interface. Non-linear least squares optimization problems are solved using the Ceres Solver library \cite{CeresSolver}. For the definition and implementation of camera models, auto-calibration and other utility functions we rely on COLMAP \cite{Schoenberger2016b}, integrated as a library. 

\subsection{Camera Calibration} \label{sec:camera_calibration}

Camera calibration is required as a preliminary step for housing calibration and therefore also included in this tool. Points in 2D image space can be related to 3D world points using the camera matrix\footnote{Throughout this work vectors and matrices are denoted as $\mat{A}$, homogeneous vectors and matrices as $\matH{B}$.} $\matH{K}$ and the camera pose $(\mat{R}|\v{t})$ up to an arbitrary scale:

\begin{equation}\label{eq:projection}
	\vH{x}\sim \matH{K}\left(\mat{R}|\v{t}\right) \vH{X}
\end{equation}

with 
\begin{equation}\label{eq:cameramat}
	\matH{K}=
	\begin{pmatrix}
		f_x & s & c_x \\
		0 & f_y & c_y \\
		0 & 0  &  1
	\end{pmatrix}
\end{equation}

Calibrating a camera now aims to recover the focal length $f_x, f_y$, principal point $(c_x, c_y)$ and skew $s$, called the intrinsic parameters. The skew is usually considered negligible for modern cameras and set to zero \cite{Hartley2004}. Estimating $\matH{K}$ can be achieved utilizing known correspondences between points in the image and world coordinates of a calibration target. For the common checkerboard calibration target the pattern corners are defined as evenly spaced, known world points. They can be automatically detected in the image yielding the required 2D-3D correspondences. Our implementation follows the approach to estimating $\matH{K}$ using a planar target described in \cite{Zhang2000}. The homographies between the target planes and the image are used to form constraints on the intrinsics, allowing a closed form solution, which is then further refined, minimizing the least squares reprojection error

\begin{equation}
	\sum_{i=1}^{n}\sum_{j=1}^{m}||\v{x}_{ij} - \v{\hat{x}}(\matH{K},\mat{R}_i,\v{t}_i,\vH{X}_j)||^2 ,
\end{equation}\label{eq:reproj_error_camera}
where the Euclidean 2D point $\v{\hat{x}}(\matH{K},\mat{R}_i,\v{t}_i,\vH{X}_j)$ is the projection of the 3D world point $\vH{X}_j$ for image $i$. 

As an extension to the intrinsic parameters in $\matH{K}$ camera models often include a term to compensate the distortion caused by camera lenses. These are defined as a 2D transformation in image space, a polynomial transforming distorted pixel coordinates into "undistored" ones. A widely used term for modeling radial distortion is \cite{Brown1971, Wei1994}:
\begin{equation}
	\begin{pmatrix} x' \\ y' \end{pmatrix} = 	\begin{pmatrix} x \left( 1 + k_1(x^2+y^2)+ k_2 (x^2+y^2)^2\right)  \\ y \left( 1 + k_1(x^2+y^2)+ k_2 (x^2+y^2)^2\right)  \end{pmatrix}
\end{equation}
Here $(x', y')$ and $(x, y)$ are the distorted and ideal normalized image coordinates respectively. There are more complex variations of this model, i.e. extending it by a term for tangential and thin-prism distortion \cite{Faig1975} or using higher order terms. During calibration these distortion parameters are initialized to zero and then jointly optimized together with $\matH{K}$. While increasing the complexity of the distortion model can improve accuracy, it can also lead to numerical instability \cite{Wei1994}. It may be sensible to start with radial distortion and only use more complex models in the presence of a larger amount of high quality calibration images. Besides polynomials, other types of parameterizations exist as well, e.g. for fisheye lenses.

In any case, the complete set of parameters defining a camera model are therefore: $\{f_x, f_y, c_x, c_y, k_1 ... k_n \}$ with the number of parameters $k$ depending on the selected distortion model. We adhere to the practice of naming certain camera models as seen in COLMAP. E.g. a camera modeled with a constant aspect ratio and a single radial distortion parameter is called "Simple Radial". For simplicity of presentation, in the subsequent sections we will only mention the matrix $\matH{K}$ for the intrinsics, but, for non-ideal lenses this must be interpreted as applying the entire intrinsics model, including distortion or fisheye parameters.

\subsection{Housing}

\begin{figure}[bt]
    \centering
    \tiny
    \def\svgwidth{0.45\columnwidth}
\begingroup%
  \makeatletter%
  \providecommand\color[2][]{%
    \errmessage{(Inkscape) Color is used for the text in Inkscape, but the package 'color.sty' is not loaded}%
    \renewcommand\color[2][]{}%
  }%
  \providecommand\transparent[1]{%
    \errmessage{(Inkscape) Transparency is used (non-zero) for the text in Inkscape, but the package 'transparent.sty' is not loaded}%
    \renewcommand\transparent[1]{}%
  }%
  \providecommand\rotatebox[2]{#2}%
  \newcommand*\fsize{\dimexpr\f@size pt\relax}%
  \newcommand*\lineheight[1]{\fontsize{\fsize}{#1\fsize}\selectfont}%
  \ifx\svgwidth\undefined%
    \setlength{\unitlength}{187.5bp}%
    \ifx\svgscale\undefined%
      \relax%
    \else%
      \setlength{\unitlength}{\unitlength * \real{\svgscale}}%
    \fi%
  \else%
    \setlength{\unitlength}{\svgwidth}%
  \fi%
  \global\let\svgwidth\undefined%
  \global\let\svgscale\undefined%
  \makeatother%
  \begin{picture}(1,0.74)%
    \lineheight{1}%
    \setlength\tabcolsep{0pt}%
    \put(0,0){\includegraphics[width=\unitlength,page=1]{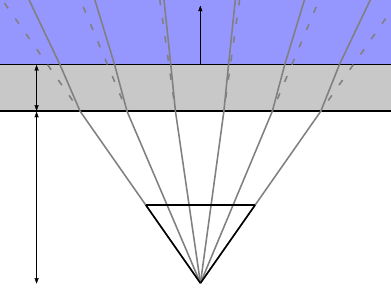}}%
    \put(0.11778976,0.50541211){\color[rgb]{0,0,0}\makebox(0,0)[lt]{\lineheight{1.25}\smash{\begin{tabular}[t]{l}$r_{flat}$\end{tabular}}}}%
    \put(0.78315472,0.35476044){\color[rgb]{0,0,0}\makebox(0,0)[lt]{\lineheight{1.25}\smash{\begin{tabular}[t]{l}$\mu_a$\end{tabular}}}}%
    \put(0.78318158,0.50087256){\color[rgb]{0,0,0}\makebox(0,0)[lt]{\lineheight{1.25}\smash{\begin{tabular}[t]{l}$\mu_g$\end{tabular}}}}%
    \put(0.78309705,0.6488704){\color[rgb]{0,0,0}\makebox(0,0)[lt]{\lineheight{1.25}\smash{\begin{tabular}[t]{l}$\mu_w$\end{tabular}}}}%
    \put(0.11788828,0.22445203){\color[rgb]{0,0,0}\makebox(0,0)[lt]{\lineheight{1.25}\smash{\begin{tabular}[t]{l}$t_{glass}$\end{tabular}}}}%
    \put(0.53174677,0.62931747){\color[rgb]{0,0,0}\makebox(0,0)[lt]{\lineheight{1.25}\smash{\begin{tabular}[t]{l}$\vec{n}$\end{tabular}}}}%
  \end{picture}%
\endgroup%

    \def\svgwidth{0.45\columnwidth}
\begingroup%
  \makeatletter%
  \providecommand\color[2][]{%
    \errmessage{(Inkscape) Color is used for the text in Inkscape, but the package 'color.sty' is not loaded}%
    \renewcommand\color[2][]{}%
  }%
  \providecommand\transparent[1]{%
    \errmessage{(Inkscape) Transparency is used (non-zero) for the text in Inkscape, but the package 'transparent.sty' is not loaded}%
    \renewcommand\transparent[1]{}%
  }%
  \providecommand\rotatebox[2]{#2}%
  \newcommand*\fsize{\dimexpr\f@size pt\relax}%
  \newcommand*\lineheight[1]{\fontsize{\fsize}{#1\fsize}\selectfont}%
  \ifx\svgwidth\undefined%
    \setlength{\unitlength}{187.5bp}%
    \ifx\svgscale\undefined%
      \relax%
    \else%
      \setlength{\unitlength}{\unitlength * \real{\svgscale}}%
    \fi%
  \else%
    \setlength{\unitlength}{\svgwidth}%
  \fi%
  \global\let\svgwidth\undefined%
  \global\let\svgscale\undefined%
  \makeatother%
  \begin{picture}(1,0.74)%
    \lineheight{1}%
    \setlength\tabcolsep{0pt}%
    \put(0,0){\includegraphics[width=\unitlength,page=1]{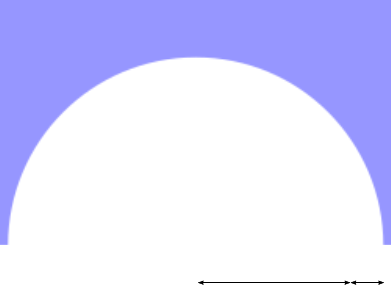}}%
    \put(0.63535602,0.05617444){\color[rgb]{0,0,0}\makebox(0,0)[lt]{\lineheight{1.25}\smash{\begin{tabular}[t]{l}$r_{dome}$\end{tabular}}}}%
    \put(0.89748132,0.05565216){\color[rgb]{0,0,0}\makebox(0,0)[lt]{\lineheight{1.25}\smash{\begin{tabular}[t]{l}$t_{glass}$\end{tabular}}}}%
    \put(0.45678113,0.13673651){\color[rgb]{0,0,0}\makebox(0,0)[lt]{\lineheight{1.25}\smash{\begin{tabular}[t]{l}$\vec{c}$\end{tabular}}}}%
    \put(0,0){\includegraphics[width=\unitlength,page=2]{domeport_diagram_mod.pdf}}%
    \put(0.71478943,0.30053705){\color[rgb]{0,0,0}\makebox(0,0)[lt]{\lineheight{1.25}\smash{\begin{tabular}[t]{l}$\mu_a$\end{tabular}}}}%
    \put(0.8145777,0.38801709){\color[rgb]{0,0,0}\makebox(0,0)[lt]{\lineheight{1.25}\smash{\begin{tabular}[t]{l}$\mu_g$\end{tabular}}}}%
    \put(0.89328967,0.46585178){\color[rgb]{0,0,0}\makebox(0,0)[lt]{\lineheight{1.25}\smash{\begin{tabular}[t]{l}$\mu_w$\end{tabular}}}}%
    \put(0,0){\includegraphics[width=\unitlength,page=3]{domeport_diagram_mod.pdf}}%
  \end{picture}%
\endgroup%

    \caption{Flat port and dome port with decentering and the resulting ray refraction.} \label{fig:housings_diagram}	
\end{figure}

For underwater applications cameras are typically protected from the environment by placing them inside a water-proof housing. The port through which the camera "looks" comes in two common shapes: A flat glass pane, a flat port, or a spherically shaped one, a dome port, c.f. figure \ref{fig:housings}, \ref{fig:housings_diagram}. The advantage of a flat port is its simplicity for manufacturing and construction. The disadvantages are that the mechanical stability against water pressure is limited, the field of view is reduced and the refraction of light inevitable. This is the advantage of a dome port, where with a perfectly aligned camera, all light rays intersect the dome at an orthogonal incident angle, avoiding refractive effects. The amount of refraction is described by Snell's Law:
\begin{equation}\label{eq:snells}
	\frac{\sin \theta_1}{\sin \theta_2} = \frac{\mu_2}{\mu_1}
\end{equation}
which relates the ratio of refractive indices $\mu_1$ and $\mu_2$, a property of the involved media, to the ratio of sines of incident and refracted ray angle $\theta_1$ and $\theta_2$. The underwater housing models effectively trace the light rays from the pinhole camera into the water. The direction of a refracted ray inside the housing glass $\vN{X}_g$ calculated from the direction of a ray in air leaving the camera $\vN{X}_a$ is \cite{Glassner1989}:
\begin{equation}\label{eq:backprojectrefract}
	\vN{X}_g = r\vN{X}_a+\vN{n}\sqrt{1-r^2 (1-(\vN{X}_a\cdot\vN{n})^2)}-r\left(\vN{X}_a\cdot\vN{n}\right),
\end{equation}
with $r=\frac{\mu_a}{\mu_g}$. 
To calculate the final ray in water, the point of intersection between the ray in air and the housing, as well as the point of intersection between the ray in the housing and water need to be found, refracting the ray at each intersection point using \eqref{eq:backprojectrefract}. The required geometric terms for ray-plane and ray-sphere intersection can be also found in \cite{Glassner1989}. 
These formulations all describe tracing a ray from the camera to world objects, i.e. back-projection. The inverse, projecting world points into the image, is a more difficult problem. For a thick flat port interface, not neglecting the housing thickness, projection requires solving a 12\textsuperscript{th} degree polynomial \cite{Agrawal2012}. 
For thick dome port housings a complete analytical solution has not been developed to our knowledge but is estimated to also be a polynomial of even higher degree \cite{She2022}. In practice this means that projection through an underwater housing is solved using a iterative numerical approach \cite{Kunz2008}. 
To calibrate the housing, camera intrinsics are considered to be known, i.e. the camera has been calibrated by itself beforehand. The required parameters to fully define a housing are the refractive indices of air, glass and water $(\mu_a, \mu_g, \mu_w)$ and the thickness of the interface $t_\mathrm{glass}$. For the flat port the interface normal $(n_x, n_y, n_z)$ and the distance from the interface to the camera projection center in normal direction $r_\mathrm{flat}$ are required. For the dome port the equivalent is the offset vector from the camera to the dome sphere center $(c_x, c_y, c_z)$ and the dome radius $r_\mathrm{dome}$. That is $\mathcal{P}_\mathrm{flat} = \{n_x, n_y, n_z,r_\mathrm{flat}, t_\mathrm{glass}, \mu_a, \mu_g, \mu_w\}$ flat port parameters and $\mathcal{P}_\mathrm{dome} = \{c_x, c_y, c_z,r_\mathrm{dome}, t_\mathrm{glass}, \mu_a, \mu_g, \mu_w\}$ dome port parameters, also shown in figure \ref{fig:housings_diagram}. 

Housing calibration is also formulated as an optimization problem minimizing reprojection errors
\begin{equation}
	\sum_{i=1}^{n}\sum_{j=1}^{m}||\v{x}_{ij} - \v{\hat{x}}_{vj}(\matH{K},\mat{R}_i,\v{t}_i, \mathcal{P},\vH{X}_j)||^2.
\end{equation}\label{eq:reproj_error_housing}
$\v{x}_{ij} - \v{\hat{x}}_{vj}(\matH{K},\mat{R}_i,\v{t}_i, \mathcal{P},\vH{X}_j)$ is the virtual camera error very similar to the one defined in \cite{Jordt2014} as modified in \cite{She2024}. 
It avoids the expensive refractive projection by calculating the error in a virtual camera space which is valid for the single ray belonging to the observation $\v{x}_{ij}$. Commonly known parts of the housing parameters, i.e. $t_\mathrm{glass}$, $r_\mathrm{dome}$ and $\mu_a$, $\mu_g$, $\mu_w$, are not optimized but rather considered constants based on the setup, as experiments show that this improves stability.
For the flat port scenario the interface normal and distance $r_\mathrm{flat}$, for the dome port the decentering $(c_x, c_y, c_z)$ are estimated. Dome port decentering is initialized using the procedure described in \cite{She2022}, where the decentering direction is derived from directly solving for the refraction center.

\subsection{Stereo Calibration}

A stereo camera system consists of two rigidly connected cameras with overlapping fields of view. Their constant relative pose allows the application of a whole group of dedicated stereo vision algorithms. Stereo calibration aims to determine this relative pose. Input for the procedure is a set of synchronized image pairs in each of which the same calibration target can be seen. Camera intrinsics can be initialized by either providing values from previous calibrations or by performing an individual calibration on the respective image set of each camera. With known intrinsics the absolute pose with respect to the calibration pattern for each camera is estimated as in single camera calibration c.f. \cite{Zhang2000}. Given the absolute pose of the first and second camera $(\mat{R}_1|\v{t}_1)$, $(\mat{R}_2|\v{t}_2)$ the relative pose of camera two with respect to one is:
\begin{align}\label{eq:relative_pose}
\left(
\begin{array}{cc} 
\mat{R}_{12}     & \v{t}_{12} \\
\v{0}^T     & 1
\end{array}
\right)
	 &= 
\left(
\begin{array}{cc}
\mat{R}_{2}     & \v{t}_{2} \\
\v{0}^T     & 1
\end{array}
\right)
  \left(
\begin{array}{cc}
\mat{R}_{1}     & \v{t}_{1} \\
\v{0}^T     & 1
\end{array}
\right)^{-1} \\
&=
  \left(
\begin{array}{cc}
\mat{R}_2 \mat{R}_1^T     & \v{t}_2 - \mat{R}_2 \mat{R}_1^T \v{t}_1 \\
\v{0}^T    & 1
\end{array}
\right) \notag
\end{align}

The relative poses determined from each image pair may differ slightly due to noisy or erroneous estimations of the absolute poses. While there are several solutions for averaging poses \cite{Hartley2013} we use a simple median relative pose. Rotations are converted to axis-angle representation and each component is sorted individually, i.e. rotation axis and translation vector dimensions as well as angle values. The median of each sorted component is then used to form an initial relative pose. This is robust against outliers of badly estimated poses. Camera intrinsics, absolute poses of the first camera and median relative pose initialize the joint optimization of a combined reprojection error of both cameras:

\begin{equation}
	\sum_{i=1}^{n}\sum_{j=1}^{m}|| e_1 ||^2 + || e_2 ||^2,
\end{equation}\label{eq:reproj_error_stereo}
with
\begin{align*}	
	e_1 &= \v{x}^{1}_{ij} - \v{\hat{x}}^{1}_{ij}(\matH{K}_1,\mat{R}_i,\v{t}_i,\vH{X}_j), \\
	e_2 &= \v{x}^{2}_{ij} - \v{\hat{x}}^{2}_{ij}(\matH{K}_2,\mat{R}_i,\v{t}_i, \mat{R}_{12}, \v{t}_{12},\vH{X}_j),
\end{align*}
being the reprojection error of the first camera, defined as the origin of the stereo pair, and second camera, positioned relatively to the first. During optimization $\matH{K}_1$ and $\matH{K}_2$ can optionally be kept constant to improve the stability when estimating the main target $(\mat{R}_{12}|\v{t}_{12})$.

\begin{figure}[t]
    \centering
    \includegraphics[height=3.4cm]{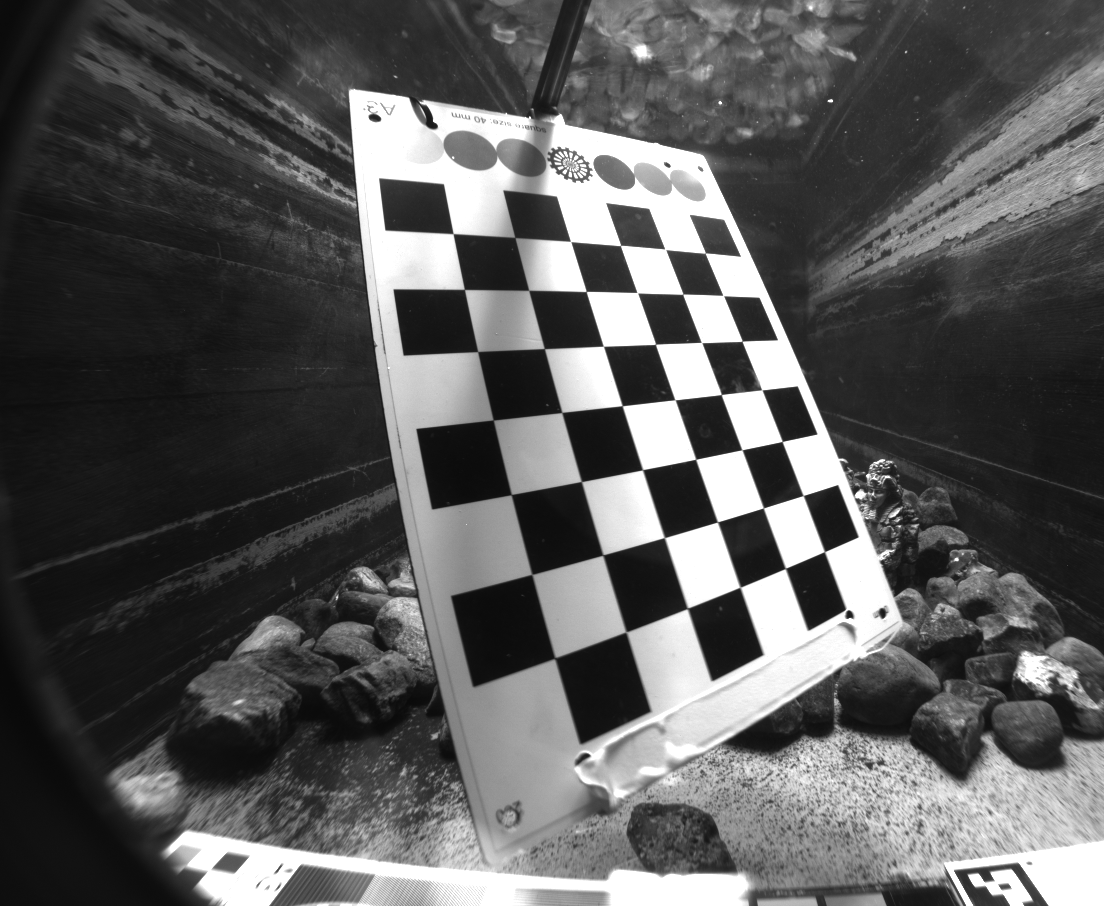}
    \includegraphics[height=3.4cm]{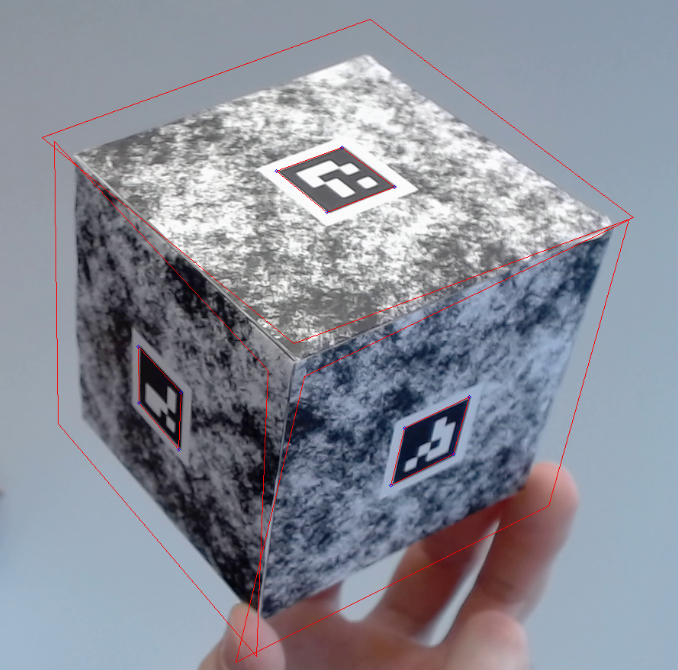}
    \\
    \vspace{0.1cm}
    \includegraphics[height=2.9cm]{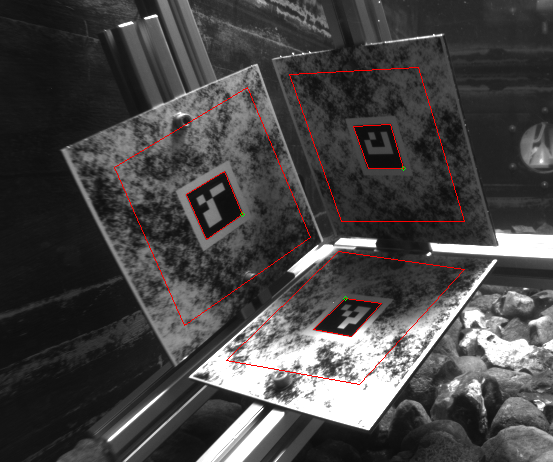}
    \includegraphics[height=2.9cm,trim={1cm 0 8cm 0},clip]{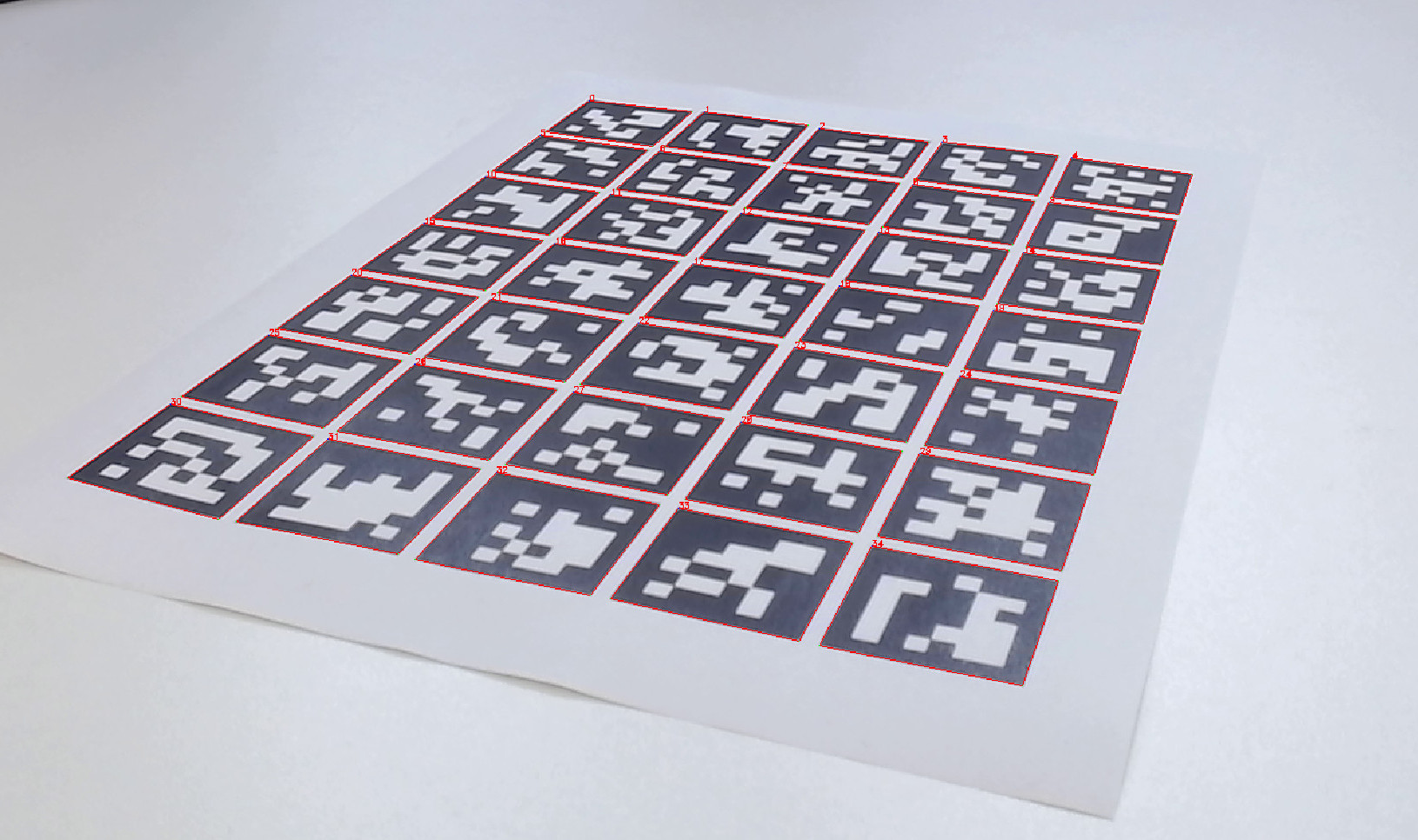}
    \caption{Possible calibration targets, checkerboard, masking auto-calibration targets and ArUco-grid board.} \label{fig:calibration_targets} 
\end{figure}

\subsection{Auto-Calibration Target}

All previously described calibration modes can be performed using the classic checkerboard calibration target. The advantage of the planar checkerboard pattern is that it is relatively easy to manufacture, a simple print can suffice for modest applications. It also allows the use of homography based constraints to solve for camera intrinsics. 
A disadvantage is that certain poses of the target do not add additional constraints to the calibration problem (e.g. pure translation is a degenerate case \cite{Zhang2000}), leading to an increased risk of (close-to) degenerate cases, and the number of feature observations is relatively low.

Employing 3D calibration targets reduces the chance for these degeneracies but such targets are more difficult to manufacture, as their 3D geometry has to be known precisely to define the correct world coordinate system for classical calibration. A third strategy is camera auto-calibration, where instead of relying on known calibration targets, the correspondences between general image features across multiple images are utilized. For an overview of auto-calibration see e.g. \cite{Hartley2004}. The approach can use arbitrary image material to calibrate a camera, which makes it very flexible, but also suffers from a few non-obvious pitfalls: Certain camera motion can lead to degenerate cases, detected features might be unreliable, scenes may contain non- rigid geometry (e.g. operator moving in the background) which must be filtered out or feature matches may not cover the whole image which can leave parts of the distortion parameters undetermined. When calibrating underwater, the requirement to move a camera around or through a feature rich scene can also be impractical. 

For application with the tool, we propose a simple masking strategy to combine the advantages of auto-calibration with the control provided by a calibration target. Instead of simply using all features in a scene as detected by e.g. a SIFT feature detector \cite{Lowe1999}, features which are not in a region of interest are masked out. The region of interest is defined by the presence of an ArUco marker \cite{GarridoJurado2014} in the image. If a marker is detected its area is enlarged naively (i.e. in pixel space, ignoring perspective) around its center by a default or user specified factor, creating a mask for features to include. This simple definition allows the usage of virtually any object as a calibration target, as long at it is rigid, has unique, SIFT detectable features and contains an ArUco marker. A practical suggestion is to generate multiple random dot patterns \cite{Li2013} designed to generate many SIFT features, add an ArUco marker to the center and attach them to an object. Examples are shown in figure \ref{fig:calibration_targets}. 
Masked, matched features are then processed in a Structure from Motion (SfM) pipeline as implemented in COLMAP \cite{Schoenberger2016b}, with an extension for refraction from \cite{She2024}. In addition to the sparse point cloud and camera poses, running SfM will also yield the auto-calibrated camera parameters. The presence of markers is additionally used to pre-calibrate the camera intrinsics. Each detected marker is interpreted as a individual view of four known 3D points in a planar coordinate system. These 2D-3D correspondences are then processed with the calibration method described in section \ref{sec:camera_calibration}. Initializing the SfM algorithm with the obtained focal length estimate improves image registration and convergence.

We display the reconstructed point cloud to visually verify a reasonable calibration result. The main advantages here are the large number of features, in the order of several thousands per image and the arbitrary geometry of the target. As long as the ArUco markers are detectable the target does not necessarily need to be planar or the texture of high quality.

\subsection{Usability Features}

Application usability is improved by a few noteworthy features.
\subsubsection*{Live Calibration}

For cameras currently connected to the machine running the tool a live calibration mode is supported. Images are acquired directly from a USB connected camera and displayed in a live view. This opens up the possibility for several user experience focused features.
\begin{itemize}
	\item The tool includes an implementation of \textit{Calibration Wizard} \cite{Peng2018} which suggests good next target poses for checkerboard-based calibration
	\item A heat-map of all previously acquired features points is shown to easily check for good coverage of detected features in all areas of the image plane
	\item A hands-free mode is available, where currently detected feature points are tracked and when kept stable automatically trigger an image acquisition
\end{itemize}

\subsubsection*{Calibration Results}

Once a calibration is completed, a concise report of all relevant parameters is shown. Overall and per view reprojection errors, the feature heat-map, estimated standard-deviations give a sense of the calibration quality. The report can be saved as text and easily parsable yaml file. Saved calibration reports can be imported to quickly re-run or initialize a new runs.

\section{Validation}

\begin{table}[b]
	\footnotesize
	\setlength{\tabcolsep}{4.3pt}
	\centering
	\caption{Rendered dataset validation} \label{tab:synthetic}
	\begin{tabular}{ r|c c c c c c }
		\hline
		\multicolumn{1}{ l|}{\textbf{Distortion}}  & $f_x$ & $c_x$ & $c_y$ & $k_1$ & $k_2$ & \\
		\hline
		\textit{True} & 1297.37 & 960 & 540 & -0.1 & -0.02 & \\
		Checker & 1302.91 & 964.91 & 535.78 & -0.010 & -0.022 &   \\
		Auto & 1295.96 & 959.46 & 539.97 & -0.10 & -0.020 &  \\
		\hline
		\multicolumn{1}{ l|}{\textbf{Dome}}  & $c_x$ & $c_y$ & $c_z$ & $||E_c||$ &  & \\						
		\hline
		\textit{True} & 0.01 & 0.006 & 0.002 & & &  \\
		Checker & 0.0103 & 0.0058 & 0.0012 & 0.0009 &  &    \\
		Auto & 0.0106 & 0.0045 & 0.0002 & 0.0024 &  &   \\
		\hline
		\multicolumn{1}{ l|}{\textbf{Flat}}  & $n_x$ & $n_y$ & $n_z$ & $r_\mathrm{flat}$ & $\epsilon$  & $||E_r||$ \\
		\hline
		\textit{True} & 0.08715 & 0 & 0.99619 & 0.02 & &  \\
		Checker & 0.0877 & -0.0025 & 0.9961 & 0.0196 & 0.15 & 0.0004 \\
		Auto & 0.0871 & 0.00005 & 0.9962 & 0.0086 & 0.005 & 0.0114 \\
		\hline
		\multicolumn{1}{ l|}{\textbf{Stereo}}  & & & &  & $\epsilon$ & $||E_t||$ \\
		\hline
		Checker & & & & & 0.04 & 0.0014 \\
		\hline
	\end{tabular}																																		
\end{table}

\begin{table}
	\small
	\setlength{\tabcolsep}{4.5pt}
	\centering
	\caption{Tank dataset validation} \label{tab:tank}
	\begin{tabular}{ r|c c c c c }
		\multicolumn{1}{ l|}{\textbf{Dome Base}}  & $c_x$ & $c_y$ & $c_z$ &  &  \\						
		\hline
		Checker & 0.0013 & 0.0010 & 0.0450 & & \\
		Auto & 0.0007 & 0.0024 & 0.0418 & & \\
		\hline
		\multicolumn{1}{ l|}{\textbf{Dome Shift}}  & $c_x$ & $c_y$ & $c_z$ & $||E_c||$ & \\						
		\hline
		Checker & 0.0099 & 0.0013 & 0.0440 & 0.0013 & \\
		Auto & 0.0084 & 0.00226 & 0.0439 & 0.002 & \\
		\hline
		\multicolumn{1}{ l|}{\textbf{Flat Base}}  & $n_x$ & $n_y$ & $n_z$ & $r_\mathrm{flat}$ & \\
		\hline
		Checker & 0.0111 & -0.0016 & 0.9999 & 0.0243 & \\
		Auto & 0.0089 & 0.0068 & 0.9999 & 0.2314 & \\
		\hline
		\multicolumn{1}{ l|}{\textbf{Flat Tilt}}  & $n_x$ & $n_y$ & $n_z$ & $r_\mathrm{flat}$ & $ \epsilon $ \\
		\hline
		Checker & 0.1021 & -0.0064 & 0.9948 & 0.0486 & 0.23 \\
		Auto & 0.0962 & 0.0032 & 0.9954 & 0.3774 & 0.01 \\
		\hline
	\end{tabular}
\end{table}

The tool implementation is validated on synthetic and real world data. The evaluation is performed in a end-to-end manner from images, in contrast to e.g. simulating numerically by projecting 3D point values. Each component, camera, housing and stereo calibration is tested individually on synthetic data, where ground truth values are known. We also test housing calibration on real world images since this is the major use case of this tool. All validation images are provided in the repository.

\subsection{Simulated Images}

Refracted images are rendered using the physically based renderer Mitsuba \cite{Mitsuba3}. For camera calibration validation the rendered images are artificially distorted using the radial distortion model. For stereo calibration intrinsics are considered to be known. For "underwater" renderings only the geometric effects of refraction are modeled, radiometric effects such as attenuation or scattering are ignored. Feature extraction algorithms for boards and SIFT include intensity normalizing steps which minimize their influence. Rendered images consist of sets for dome and flat port underwater as well as an "in-air" set to validate lens distortion. Each of them are generated for checkerboard and auto-calibration targets individually. 
The dome port is simulated with 5 cm radius and 6 mm thickness, the flat port with 2 cm distance and 14 mm thickness. For both, $(\mu_a, \mu_g, \mu_w) = \mathrm{(1, 1.473, 1.334)}$. The underlying camera has a field of view of 73°, which corresponds to a focal length of 1297.3655, with a resolution of 1920x1080 and a principal point in the image center. For the auto-calibration target the dome decentering is initialized with $(0,0,0)$, the flat port normal with $(0,0,1)$ and double the ground truth interface distance (the distance of interface to camera projection center is usually not known exactly). All estimated values with the respective simulation ground truth is given in table \ref{tab:synthetic}. Note that the ground truth interface normal represents a rotation around the y-axis by 5°. For the dome port $||E_c||$ denotes the dome port decentering estimation error in m. For the flat port $||E_r||$ gives the error in interface distance estimation in m and $\epsilon$ the angular error for normal estimation in degrees. Stereo calibration is currently only implemented for the checker board target. All rendered image sets consist of 25 images each.

\subsection{Tank Images}

\begin{figure}
	\centering
	\def\svgwidth{0.89\columnwidth}
\begingroup%
  \makeatletter%
  \providecommand\color[2][]{%
    \errmessage{(Inkscape) Color is used for the text in Inkscape, but the package 'color.sty' is not loaded}%
    \renewcommand\color[2][]{}%
  }%
  \providecommand\transparent[1]{%
    \errmessage{(Inkscape) Transparency is used (non-zero) for the text in Inkscape, but the package 'transparent.sty' is not loaded}%
    \renewcommand\transparent[1]{}%
  }%
  \providecommand\rotatebox[2]{#2}%
  \newcommand*\fsize{\dimexpr\f@size pt\relax}%
  \newcommand*\lineheight[1]{\fontsize{\fsize}{#1\fsize}\selectfont}%
  \ifx\svgwidth\undefined%
    \setlength{\unitlength}{1650bp}%
    \ifx\svgscale\undefined%
      \relax%
    \else%
      \setlength{\unitlength}{\unitlength * \real{\svgscale}}%
    \fi%
  \else%
    \setlength{\unitlength}{\svgwidth}%
  \fi%
  \global\let\svgwidth\undefined%
  \global\let\svgscale\undefined%
  \makeatother%
  \begin{picture}(1,0.63636364)%
    \lineheight{1}%
    \setlength\tabcolsep{0pt}%
    \put(0,0){\includegraphics[width=\unitlength,page=1]{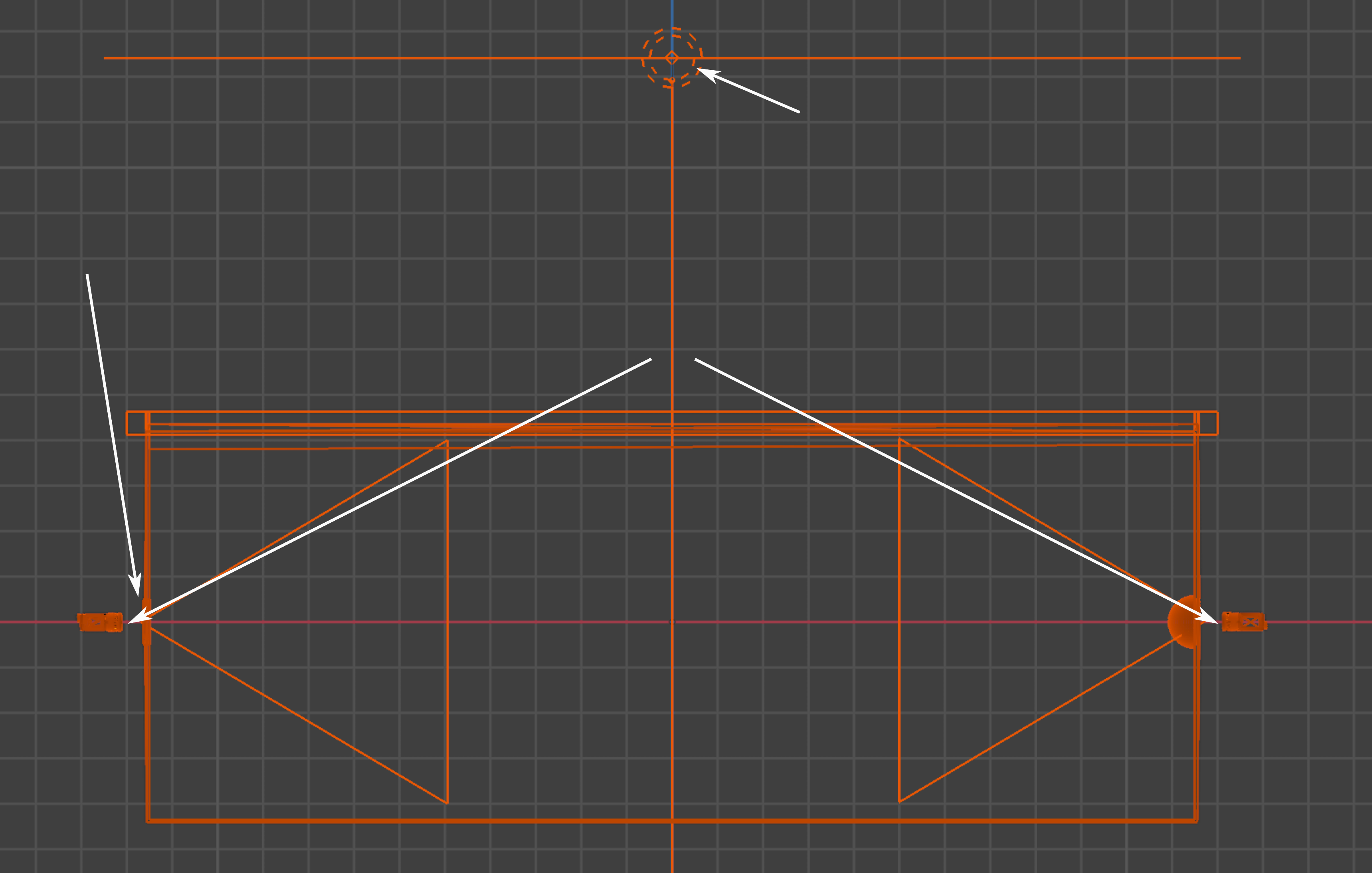}}%
    \put(0.02192848,0.45092288){\color[rgb]{1,1,1}\makebox(0,0)[lt]{\lineheight{1.25}\smash{\begin{tabular}[t]{l}Flat Port\end{tabular}}}}%
    \put(0.80095908,0.45081608){\color[rgb]{1,1,1}\makebox(0,0)[lt]{\lineheight{1.25}\smash{\begin{tabular}[t]{l}Dome Port\end{tabular}}}}%
    \put(0.58714378,0.539306){\color[rgb]{1,1,1}\makebox(0,0)[lt]{\lineheight{1.25}\smash{\begin{tabular}[t]{l}Lighting\end{tabular}}}}%
    \put(0.43091242,0.3848723){\color[rgb]{1,1,1}\makebox(0,0)[lt]{\lineheight{1.25}\smash{\begin{tabular}[t]{l}Camera\end{tabular}}}}%
    \put(0,0){\includegraphics[width=\unitlength,page=2]{tank_schema.pdf}}%
  \end{picture}%
\endgroup%

	\vspace{0.12cm}
	
	\includegraphics[height=3.1cm]{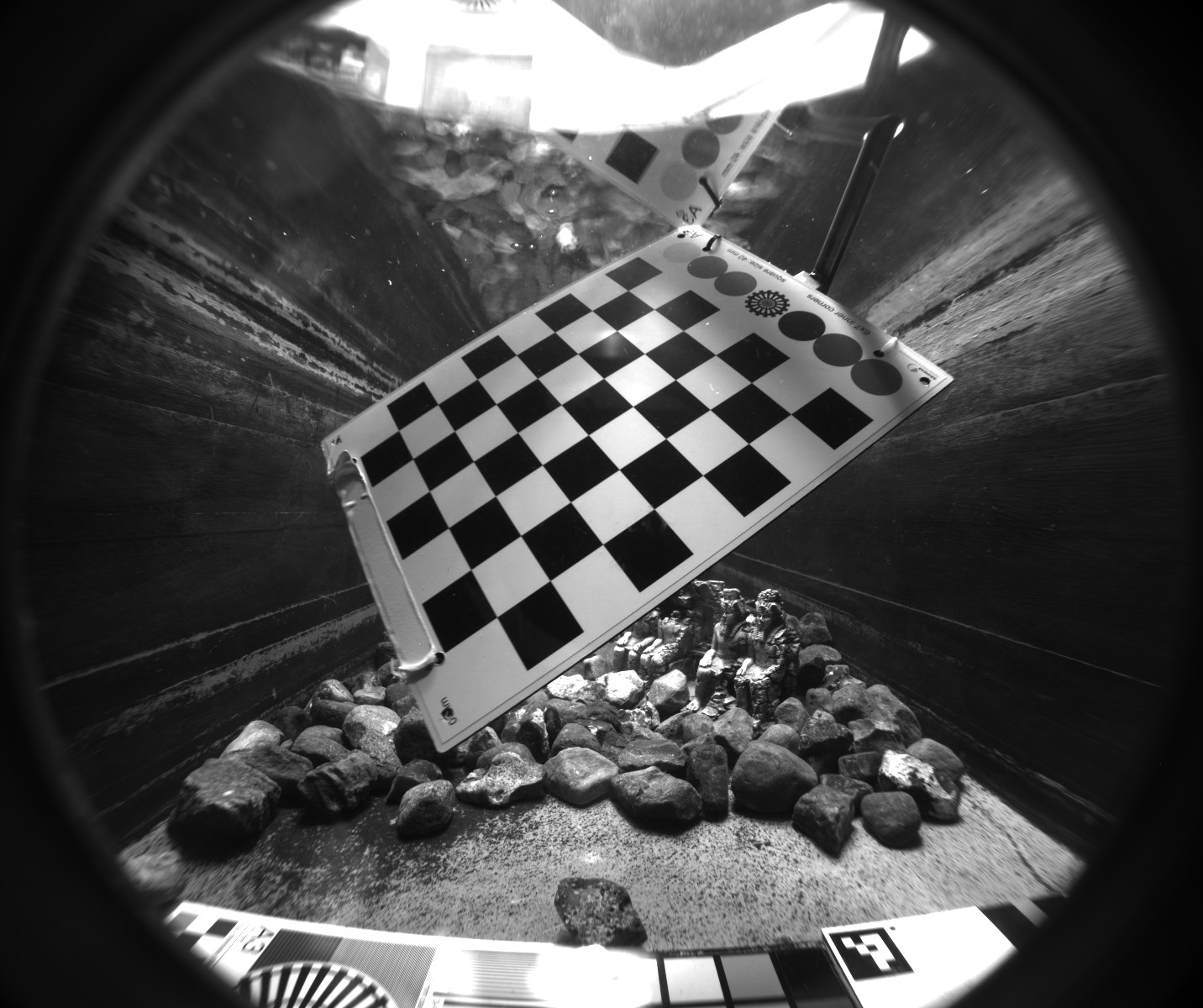}
	\includegraphics[height=3.1cm]{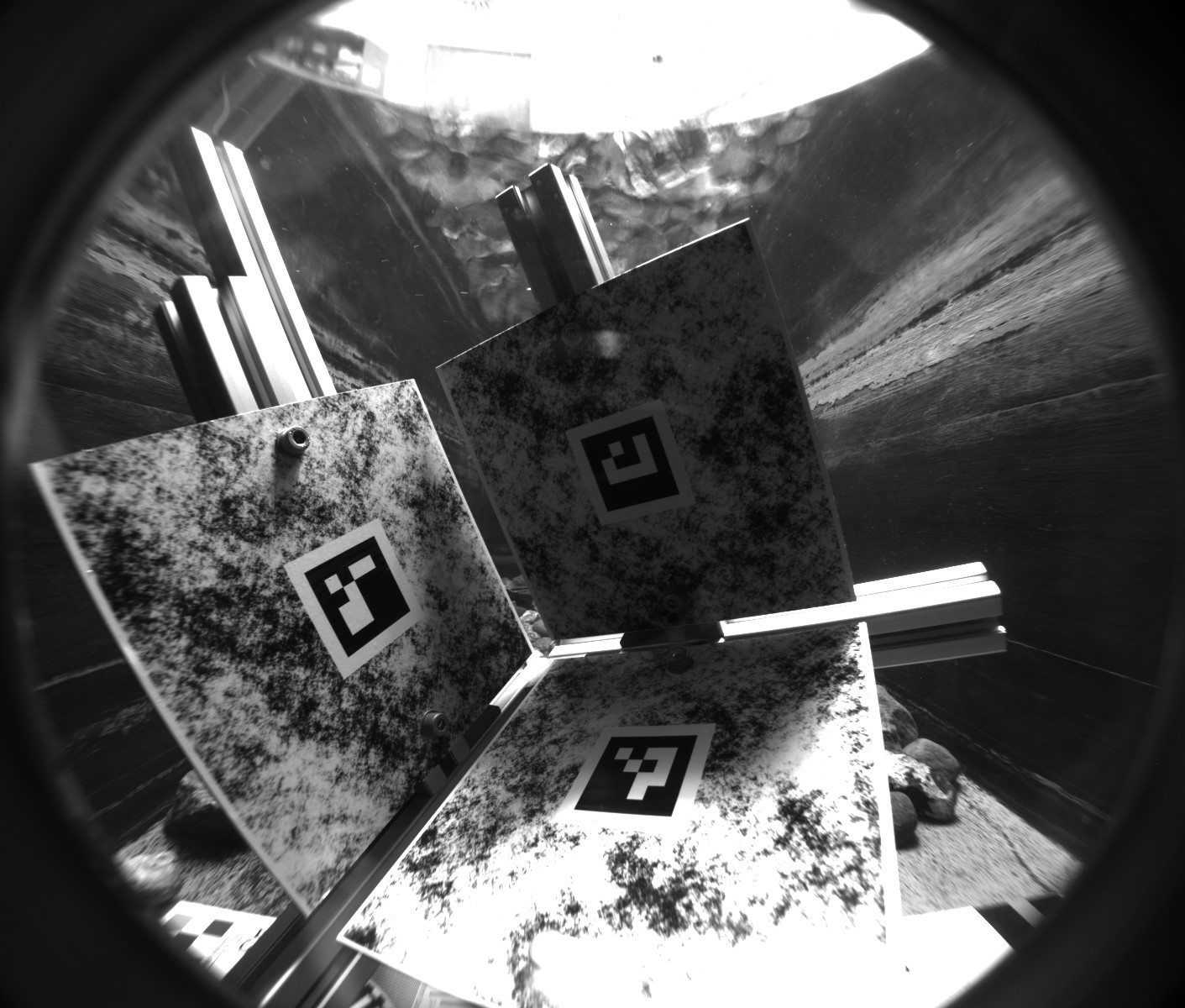}
	\caption{Schematic of the tank setup (\textit{top}) and tank image samples for checkerboard and auto-calibration target behind the dome (\textit{bottom}). Tank schematic adapted from \cite{Nakath2022}.} \label{fig:tank_images}
\end{figure}

Housing calibration is verified from real images. They are taken looking through the ports on a lab tank, as illustrated in figure \ref{fig:tank_images}. Because mechanically measuring the absolute  projection center inside the dome or flat port as a ground truth reference is very challenging we instead opted for a "differential" (relative motion) experiment. After acquiring a first set of images the camera was moved inside the dome or tilted in front of the flat port by a controlled movement. For the dome port the camera was shifted by 10 mm in between sets, in the flat port setup the camera was tilted by 5°. Table \ref{tab:tank} shows the estimated values for each target and set, as well as the dome displacement error $||E_c||$ in m and the interface normal angular error $\epsilon$ in degrees, as the deviation from the expected value for each target.

The camera used is a  Basler acA1920-50gm. Because of the optics vignette the effective resolution is 1480 x 1200, with a field of view of approx. 92°. The dome is 7 mm thick, with a radius of 50mm, made from Vitrovex glass with $\mu_g = 1.473$. The flat port is a 10 mm thick PMMA pane with $\mu_g = 1.49$. The interface normal is initialized as aligned, with a distance of estimated 1 cm. The dome decentering is initialized with 4 cm decentering in z-direction, estimated from the physical setup. The camera is calibrated in air using checkerboard images for the checkerboard set and auto-calibrated for the auto-calibration set employing a distortion model with two radial and two tangential distortion parameters. The resulting calibration parameters are available in the provided data set.

\subsection{Discussion}

Both the checkerboard and auto calibration can closely recover camera intrinsics in the synthetic images. Bigger differences are only visible in the housing calibration sets. The checkerboard calibration estimations of dome port decentering are correct to millimeter precision, which roughly matches the accuracy reported in \cite{She2022}. Interface distance (only known for the synthetic set) and tilt are well estimated using the checkerboard (with known absolute 3D coordinates) as well. The remaining error is similar to the ones reported in \cite{Agrawal2012, JordtSedlazeck2012, Elnashef2022}. 
Note that for auto-calibration (unknown 3D target), absolute distances can not be obtained as the scene scale is unknown. Even a perfectly known interface thickness is - in practice -  a too weak constraint to exactly determine the other parameters absolutely. This has already been noted in \cite{Elnashef2022} and \cite{She2022}. The refractive 3D reconstructions will often show scales which do not exactly match the interface thickness, in case no other absolute 3D information is present. This problem could be tackled by including scene scale constraints into the auto-calibration (i.e. some known properties of 3D target). One obvious source is the markers on the target, for which a user provided scale could be used to restrict the optimization.

\section{CONCLUSION}

In this work we present a stand-alone, cross-platform application with the capability to calibrate cameras, stereo setups and underwater housings. The main target is the calibration for computer vision in underwater scenarios, but standard calibration is supported as well. It can recover camera intrinsics, distortion parameters, stereo pose, flat port interface normal and distance as well as dome port decentering. A newly proposed auto-calibration target is supported which allows the explicit masking of undesired features combining the advantages of a calibration target with the effectiveness of auto-calibration. The implementation is verified on rendered and real images, which we release together with the source code and application. For future work we will allow entering scale constraints from the auto-calibration target to further improve its performance for housing calibration.





\bibliographystyle{IEEEtran}
\bibliography{IEEEabrv,calibmar.bib}

\begin{thebibliography}{10}
\providecommand{\url}[1]{#1}
\csname url@samestyle\endcsname
\providecommand{\newblock}{\relax}
\providecommand{\bibinfo}[2]{#2}
\providecommand{\BIBentrySTDinterwordspacing}{\spaceskip=0pt\relax}
\providecommand{\BIBentryALTinterwordstretchfactor}{4}
\providecommand{\BIBentryALTinterwordspacing}{\spaceskip=\fontdimen2\font plus
\BIBentryALTinterwordstretchfactor\fontdimen3\font minus
  \fontdimen4\font\relax}
\providecommand{\BIBforeignlanguage}[2]{{%
\expandafter\ifx\csname l@#1\endcsname\relax
\typeout{** WARNING: IEEEtran.bst: No hyphenation pattern has been}%
\typeout{** loaded for the language `#1'. Using the pattern for}%
\typeout{** the default language instead.}%
\else
\language=\csname l@#1\endcsname
\fi
#2}}
\providecommand{\BIBdecl}{\relax}
\BIBdecl

\bibitem{Lavest2000}
J.~M. Lavest, G.~Rives, and J.~T. Laprest{\'{e}}, ``Underwater camera
  calibration,'' in \emph{Lecture Notes in Computer Science}.\hskip 1em plus
  0.5em minus 0.4em\relax Springer Berlin Heidelberg, 2000, pp. 654--668.

\bibitem{Shortis2015}
M.~Shortis, ``Calibration techniques for accurate measurements by underwater
  camera systems,'' \emph{Sensors}, vol.~15, no.~12, pp. 30\,810--30\,826, Dec.
  2015.

\bibitem{Sedlazeck2012}
A.~Sedlazeck and R.~Koch, \emph{Perspective and Non-perspective Camera Models
  in Underwater Imaging – Overview and Error Analysis}.\hskip 1em plus 0.5em
  minus 0.4em\relax Springer Berlin Heidelberg, 2012, pp. 212--242.

\bibitem{Treibitz2012}
T.~Treibitz, Y.~Schechner, C.~Kunz, and H.~Singh, ``Flat refractive geometry,''
  \emph{{IEEE} Transactions on Pattern Analysis and Machine Intelligence},
  vol.~34, no.~1, pp. 51--65, jan 2012.

\bibitem{kahmen2020impact}
O.~Kahmen, R.~Rofallski, and T.~Luhmann, ``Impact of stereo camera calibration
  to object accuracy in multimedia photogrammetry,'' \emph{Remote Sensing},
  vol.~12, no.~12, p. 2057, 2020.

\bibitem{Luczynski2017}
T.~{\L}uczy{\'{n}}ski, M.~Pfingsthorn, and A.~Birk, ``The pinax-model for
  accurate and efficient refraction correction of underwater cameras in
  flat-pane housings,'' \emph{Ocean Engineering}, vol. 133, pp. 9--22, mar
  2017.

\bibitem{nocerino2021bundle}
E.~Nocerino, F.~Menna, and A.~Gr{\"u}n, ``Bundle adjustment with polynomial
  point-to-camera distance dependent corrections for underwater
  photogrammetry,'' \emph{The International Archives of the Photogrammetry,
  Remote Sensing and Spatial Information Sciences}, vol.~43, pp. 673--679,
  2021.

\bibitem{menna2017optical}
F.~Menna, E.~Nocerino, and F.~Remondino, ``Optical aberrations in underwater
  photogrammetry with flat and hemispherical dome ports,'' in
  \emph{Videometrics, Range Imaging, and Applications XIV}, vol. 10332.\hskip
  1em plus 0.5em minus 0.4em\relax SPIE, 2017, pp. 28--41.

\bibitem{She2022}
M.~She, D.~Nakath, Y.~Song, and K.~Köser, ``Refractive geometry for underwater
  domes,'' \emph{ISPRS Journal of Photogrammetry and Remote Sensing}, vol. 183,
  pp. 525--540, 2022.

\bibitem{kotowski1988phototriangulation}
R.~Kotowski, ``Phototriangulation in multi-media photogrammetry,''
  \emph{International Archives of Photogrammetry and Remote Sensing}, vol.~27,
  no.~B5, pp. 324--334, 1988.

\bibitem{Glassner1989}
A.~S. Glassner, Ed., \emph{An Introduction to Ray Tracing}.\hskip 1em plus
  0.5em minus 0.4em\relax Academic Press Ltd., 1989.

\bibitem{maas1992digitale}
H.-G. Maas, ``Digitale photogrammetrie in der dreidimensionalen
  str{\"o}mungsmesstechnik,'' Ph.D. dissertation, ETH Zurich, 1992.

\bibitem{JordtSedlazeck2012}
A.~Jordt-Sedlazeck and R.~Koch, ``Refractive calibration of underwater
  cameras,'' in \emph{Computer Vision {\textendash} {ECCV} 2012}.\hskip 1em
  plus 0.5em minus 0.4em\relax Springer Berlin Heidelberg, 2012, pp. 846--859.

\bibitem{Agrawal2012}
A.~Agrawal, S.~Ramalingam, Y.~Taguchi, and V.~Chari, ``A theory of multi-layer
  flat refractive geometry,'' in \emph{2012 {IEEE} Conference on Computer
  Vision and Pattern Recognition}.\hskip 1em plus 0.5em minus 0.4em\relax
  {IEEE}, jun 2012.

\bibitem{elnashef2023theory}
B.~Elnashef and S.~Filin, ``Theory and closed-form solutions for three-and
  n-layer flat refractive geometry,'' \emph{International Journal of Computer
  Vision}, vol. 131, no.~4, pp. 877--898, 2023.

\bibitem{she2019adjustment}
M.~She, Y.~Song, J.~Mohrmann, and K.~K{\"o}ser, ``Adjustment and calibration of
  dome port camera systems for underwater vision,'' in \emph{Pattern
  Recognition: 41st DAGM German Conference, DAGM GCPR 2019, Dortmund, Germany,
  September 10--13, 2019, Proceedings 41}.\hskip 1em plus 0.5em minus
  0.4em\relax Springer, 2019, pp. 79--92.

\bibitem{Kunz2008}
C.~Kunz and H.~Singh, ``Hemispherical refraction and camera calibration in
  underwater vision,'' in \emph{{OCEANS} 2008}.\hskip 1em plus 0.5em minus
  0.4em\relax {IEEE}, 2008.

\bibitem{JordtSedlazeck2013}
A.~Jordt-Sedlazeck and R.~Koch, ``Refractive structure-from-motion on
  underwater images,'' in \emph{2013 {IEEE} International Conference on
  Computer Vision}.\hskip 1em plus 0.5em minus 0.4em\relax {IEEE}, dec 2013.

\bibitem{jordt2016refractive}
A.~Jordt, K.~K{\"o}ser, and R.~Koch, ``Refractive 3d reconstruction on
  underwater images,'' \emph{Methods in Oceanography}, vol.~15, pp. 90--113,
  2016.

\bibitem{Chadebecq2019}
F.~Chadebecq, F.~Vasconcelos, R.~Lacher, E.~Maneas, A.~Desjardins, S.~Ourselin,
  T.~Vercauteren, and D.~Stoyanov, ``Refractive two-view reconstruction for
  underwater 3d vision,'' \emph{International Journal of Computer Vision}, vol.
  128, no.~5, pp. 1101--1117, nov 2019.

\bibitem{Kang2017}
L.~Kang, L.~Wu, Y.~Wei, S.~Lao, and Y.-H. Yang, ``Two-view underwater 3d
  reconstruction for cameras with unknown poses under flat refractive
  interfaces,'' \emph{Pattern Recognition}, vol.~69, pp. 251--269, sep 2017.

\bibitem{Kang2012}
L.~Kang, L.~Wu, and Y.-H. Yang, ``Two-view underwater structure and motion for
  cameras under flat refractive interfaces,'' in \emph{Computer Vision
  {\textendash} {ECCV} 2012}.\hskip 1em plus 0.5em minus 0.4em\relax Springer
  Berlin Heidelberg, 2012, pp. 303--316.

\bibitem{Seegraeber2023}
F.~Seegräber, P.~Schöntag, F.~Woelk, and K.~Köser, ``Underwater multiview
  stereo using axial camera models,'' in \emph{DAGM GCPR 2023}, 2023.

\bibitem{Elnashef2023}
B.~Elnashef and S.~Filin, ``Drift reduction in underwater egomotion computation
  by axial camera modeling,'' \emph{IEEE Robotics and Automation Letters},
  vol.~8, no.~5, pp. 2676--2683, May 2023.

\bibitem{Peng2018}
S.~Peng and P.~F. Sturm, ``Calibration wizard: A guidance system for camera
  calibration based on modelling geometric and corner uncertainty,'' \emph{2019
  IEEE/CVF International Conference on Computer Vision (ICCV)}, pp. 1497--1505,
  2018.

\bibitem{Bouguet2022}
J.-Y. Bouguet, ``\BIBforeignlanguage{en}{Camera calibration toolbox for
  matlab},'' 2022.

\bibitem{OpenCV2024}
\BIBentryALTinterwordspacing
OpenCV, ``{The OpenCV Library},'' 2024. [Online]. Available:
  \url{https://opencv.org/}
\BIBentrySTDinterwordspacing

\bibitem{Furgale2013}
P.~Furgale, J.~Rehder, and R.~Siegwart, ``Unified temporal and spatial
  calibration for multi-sensor systems,'' in \emph{2013 IEEE/RSJ International
  Conference on Intelligent Robots and Systems}.\hskip 1em plus 0.5em minus
  0.4em\relax IEEE, Nov. 2013.

\bibitem{Li2013}
B.~Li, L.~Heng, K.~Koser, and M.~Pollefeys, ``A multiple-camera system
  calibration toolbox using a feature descriptor-based calibration pattern,''
  in \emph{2013 IEEE/RSJ International Conference on Intelligent Robots and
  Systems}.\hskip 1em plus 0.5em minus 0.4em\relax IEEE, Nov. 2013.

\bibitem{Scaramuzza2006}
D.~Scaramuzza, A.~Martinelli, and R.~Siegwart, ``A toolbox for easily
  calibrating omnidirectional cameras,'' in \emph{2006 IEEE/RSJ International
  Conference on Intelligent Robots and Systems}.\hskip 1em plus 0.5em minus
  0.4em\relax IEEE, Oct. 2006.

\bibitem{Schramm2021}
S.~Schramm, J.~Rangel, D.~A. Salazar, R.~Schmoll, and A.~Kroll, ``Target
  analysis for the multispectral geometric calibration of cameras in visual and
  infrared spectral range,'' \emph{IEEE Sensors Journal}, vol.~21, no.~2, pp.
  2159--2168, Jan. 2021.

\bibitem{CeresSolver}
\BIBentryALTinterwordspacing
S.~Agarwal, K.~Mierle, and T.~C.~S. Team, ``{Ceres Solver},'' 10 2023.
  [Online]. Available: \url{https://github.com/ceres-solver/ceres-solver}
\BIBentrySTDinterwordspacing

\bibitem{Schoenberger2016b}
J.~L. Schonberger and J.-M. Frahm, ``Structure-from-motion revisited,'' in
  \emph{2016 IEEE Conference on Computer Vision and Pattern Recognition
  (CVPR)}.\hskip 1em plus 0.5em minus 0.4em\relax IEEE, Jun. 2016.

\bibitem{Hartley2004}
R.~I. Hartley and A.~Zisserman, \emph{Multiple View Geometry in Computer
  Vision}, 2nd~ed.\hskip 1em plus 0.5em minus 0.4em\relax Cambridge University
  Press, ISBN: 0521540518, 2004.

\bibitem{Zhang2000}
Z.~Zhang, ``A flexible new technique for camera calibration,'' \emph{IEEE
  Transactions on Pattern Analysis and Machine Intelligence}, vol.~22, no.~11,
  pp. 1330--1334, 2000.

\bibitem{Brown1971}
D.~Brown, ``Close-range camera calibration,'' 1971.

\bibitem{Wei1994}
G.-Q. Wei and S.~D. Ma, ``Implicit and explicit camera calibration: theory and
  experiments,'' \emph{IEEE Transactions on Pattern Analysis and Machine
  Intelligence}, vol.~16, no.~5, pp. 469--480, May 1994.

\bibitem{Faig1975}
W.~Faig, ``Calibration of close-range photogrammetric systems: Mathematical
  formulation,'' \emph{Photogrammetric Engineering and Remote Sensing},
  vol.~41, 1975.

\bibitem{Jordt2014}
A.~Jordt, ``Underwater 3d reconstruction based on physical models for
  refraction and underwater light propagation,'' 2014.

\bibitem{She2024}
M.~She, F.~Seegräber, D.~Nakath, and K.~Köser, ``Refractive colmap:
  Refractive structure-from-motion revisited,'' in \emph{2024 IEEE/RSJ
  International Conference on Intelligent Robots and Systems (IROS)}, 2024, pp.
  12\,816--12\,823.

\bibitem{Hartley2013}
R.~Hartley, J.~Trumpf, Y.~Dai, and H.~Li, ``Rotation averaging,''
  \emph{International Journal of Computer Vision}, vol. 103, no.~3, pp.
  267--305, Jan. 2013.

\bibitem{Lowe1999}
D.~Lowe, ``Object recognition from local scale-invariant features,'' in
  \emph{Proceedings of the Seventh IEEE International Conference on Computer
  Vision}.\hskip 1em plus 0.5em minus 0.4em\relax IEEE, 1999.

\bibitem{GarridoJurado2014}
S.~Garrido-Jurado, R.~Muñoz-Salinas, F.~Madrid-Cuevas, and M.~Marín-Jiménez,
  ``Automatic generation and detection of highly reliable fiducial markers
  under occlusion,'' \emph{Pattern Recognition}, vol.~47, no.~6, pp.
  2280--2292, Jun. 2014.

\bibitem{Mitsuba3}
W.~Jakob, S.~Speierer, N.~Roussel, M.~Nimier-David, D.~Vicini, T.~Zeltner,
  B.~Nicolet, M.~Crespo, V.~Leroy, and Z.~Zhang, ``Mitsuba 3 renderer,'' 2022,
  https://mitsuba-renderer.org.

\bibitem{Nakath2022}
D.~Nakath, M.~She, Y.~Song, and K.~K{\"o}ser, ``An optical digital twin for
  underwater photogrammetry,'' \emph{PFG -- Journal of Photogrammetry, Remote
  Sensing and Geoinformation Science}, Jan 2022.

\bibitem{Elnashef2022}
B.~Elnashef and S.~Filin, ``Target-free calibration of flat refractive imaging
  systems using two-view geometry,'' \emph{Optics and Lasers in Engineering},
  vol. 150, p. 106856, Mar. 2022.

\end{thebibliography}

\end{document}